\documentclass[runningheads]{llncs}
\usepackage{graphicx}
\newcommand*\samethanks[1][\value{footnote}]{\footnotemark[#1]}
\usepackage[table]{xcolor}
\usepackage{chngpage}
\usepackage{float}
\usepackage{etoolbox}
\apptocmd{\sloppy}{\hbadness 10000\relax}{}{}

\begin{document}

\title{FedCostWAvg: A new averaging for better Federated Learning}

\author{Leon M\"achler\inst{6} \and
Ivan Ezhov\inst{1,3} \and
Florian Kofler\inst{1,2,3} \and
Suprosanna Shit\inst{1,3} \and
Johannes C. Paetzold\inst{1,3,5} \and
Timo Loehr \inst{1,3} \and
Claus Zimmer\inst{2} \and
Benedikt Wiestler\inst{2} \and
Bjoern H. Menze\inst{1,3,4}\samethanks}



\authorrunning{M\"achler et al.}
%
\institute{Department of Informatics, Technical University Munich, Germany\and 
Department of Diagnostic and Interventional Neuroradiology, School of Medicine, Klinikum rechts der Isar, Technical University of Munich, Germany\and
TranslaTUM - Central Institute for Translational Cancer Research, Technical University of Munich, Germany\and
Department of Quantitative Biomedicine, University of Zurich, Switzerland \and
ITERM Institute Helmholtz Zentrum Muenchen, Neuherberg, Germany \and
École Normale Supérieure, Paris, France \\
\email{leon-philipp.machler@ens.fr}\\
}


%

\maketitle
\setcounter{footnote}{0}
\begin{abstract}
We propose a simple new aggregation strategy for federated learning that won the MICCAI Federated Tumor Segmentation Challenge 2021 (FETS), the first ever challenge on Federated Learning in the Machine Learning community. Our method addresses the problem of how to aggregate multiple models that were trained on different data sets. Conceptually, we propose a new way to choose the weights when averaging the different models, thereby extending the current state of the art (FedAvg). Empirical validation demonstrates that our approach reaches a notable improvement in segmentation performance compared to FedAvg.
\end{abstract}

\section{Introduction}
\subsection{Motivation}
Preserving data privacy is of paramount importance for confidentiality-critical fields such as the medical domain. Today it is not uncommon that large volumes of private medical records are illegally released to the \emph{dark web}\cite{meddatahack}. To prevent such incidents, often large amounts of resources are allocated but cannot guarantee full security. Among many precautions, reducing human (including IT specialists) exposure to the data is highly desirable to reduce the chance of compromising data protection by human failure. 

\begin{figure}[ht]
    \label{fl}
    \centering
    \includegraphics[width=\textwidth]{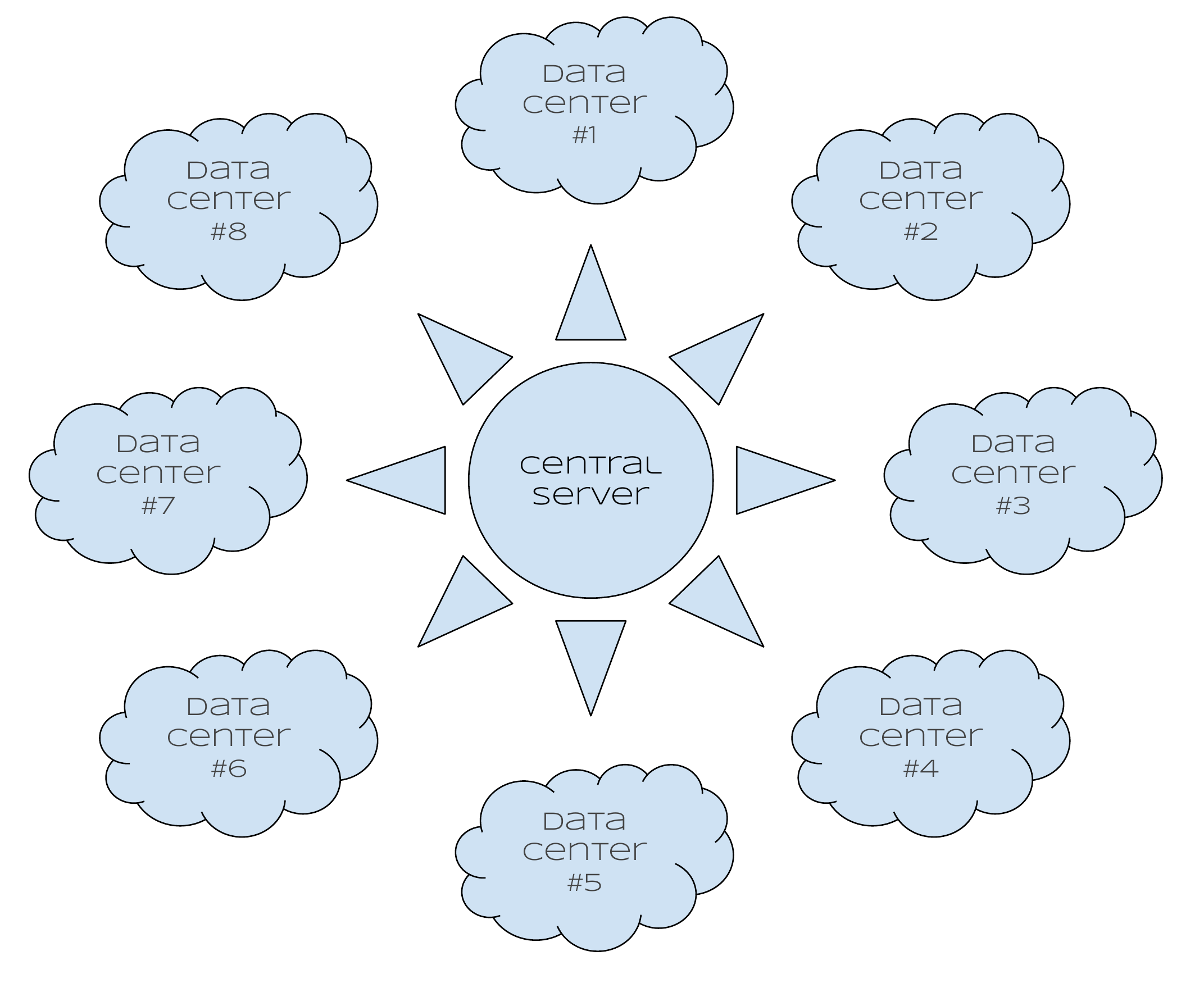}
    \caption{Schematic illustration of the federated learning concept. Within multiple data centers, a model is trained for our task. Next, parameters are sent to the central server, where aggregation of the parameters takes place. An aggregated global configuration of the parameters is broadcasted back to the centers. The procedure repeats until convergence or some other limit is reached.}
    \label{fig2}
\end{figure} 
\subsection{The typical training scenario}
In machine learning, a common scenario today looks like this: One or more institutions (companies, research institutes, governments, etc.) gather data, share it with data scientists who, in turn, train some sort of a model using the data. For example, a group of hospitals share MRI scans of tumors with the medical community to help with the development of an automatic tumor segmentation model. One problem with this approach is that the data, once it is shared, might get leaked, misused, or stolen from the developers. Other hurdles include legal reasons that might make it impossible for the hospitals to share and pool the data in the first place.  

\subsection{Federated Learning}
Conventional machine learning requires exposing training data to a learning algorithm and its developers. When several data sources are involved, the pooling together of the data to create a single data set is also required. New approaches like \emph{Federated learning} (FL) \cite{rieke2020future} allow to separate model training from developer access while also not requiring any pooling of data. FL was introduced in a series of seminal works starting from 2015 \cite{KonecnyMYRSB16,KonecnyMR15,mcmahan2017communication}. FL is a protocol consisting of two alternating steps: a) independent training of models on local entities with their respective unique corpus of data, and b) broadcasting back of only the weights of the trained models to a central entity where the weights are aggregated and a new model is redistributed. The choice of which type of model or network to perform step (a) is dictated by the task (e.g., classification, segmentation, etc.) and can be made based on the state-of-the-art in the respective task. The new FL scenario looks like this: A developer sends his or her model to all the institutions that own training data, the institutions locally train the model for the developer and send the newly trained models back. In this way, the developer can train their model while never getting any access to the data. In this setting however a new problem arises. 

\subsection{The aggregation problem}
How to aggregate the different models that come back? A naive approach to solve the problem would be: \begin{enumerate}
	\item Send an initial model to the first data center
	\item Get back a newly trained model and send it to the second data center
	\item Repeat until all data centers have trained the model once 
\end{enumerate} Approaches like this are called sequential learning and fail due to a phenomenon called "catastrophic forgetting"~\cite{MCCLOSKEY1989109}. Effectively what would happen is that the final model would only be trained on the data of the last center and would not have generalized to the entire corpus of data. It would simply forget what was learned in the previous center as soon as it gets trained by the next. The state-of-the-art approach tries to avoid this phenomenon by including feedback from every center in each update.

\subsection{State of the art}
The seminal work of learning deep networks from decentralized data \cite{mcmahan2017communication} proposed as a solution a plain coordinate-wise mean averaging (FedAvg) of the model weights coming separately from multiple centers. Recently \cite{yurochkin2019bayesian} proposed a valuable extension to FedAvg, which takes invariance of network weights to permutations into account. In \cite{sahu2018convergence} (FedProx), the authors adjust the training loss of a local model to enforce closeness of local and global model updates. Despite methodological advances, there is neither theoretical nor practical evidence for the right recipe when choosing an aggregation strategy. In this paper, we propose a new idea on how to do aggregation.
Similar to other initiatives \cite{sekuboyina2020verse,payette2020comparison,paetzold2021whole,bilic2019liver}, the FETS challenge\footnote{https://fets-ai.github.io/Challenge/} \cite{pati2021federated} is organized to benchmark different weight aggregation strategies on the clinically important glioma segmentation problem \cite{bakas2017advancing,pati2021federated,reina2021openfl,sheller2020federated,kofler2020brats}. We contribute to the initiative by proposing an effective extension to the FedAvg strategy. When compared with the other submissions, our model significantly outperformed all of them and won the challenge. On top of that we tested the model locally on a smaller corpus of data to compare it to FedAvg. It notably improves performance compared with FedAvg at no additional compute time.
\section{Methodology}


\subsection{Segmentation network}
\begin{figure}[t!]
    \centering
    \includegraphics[width=\textwidth]{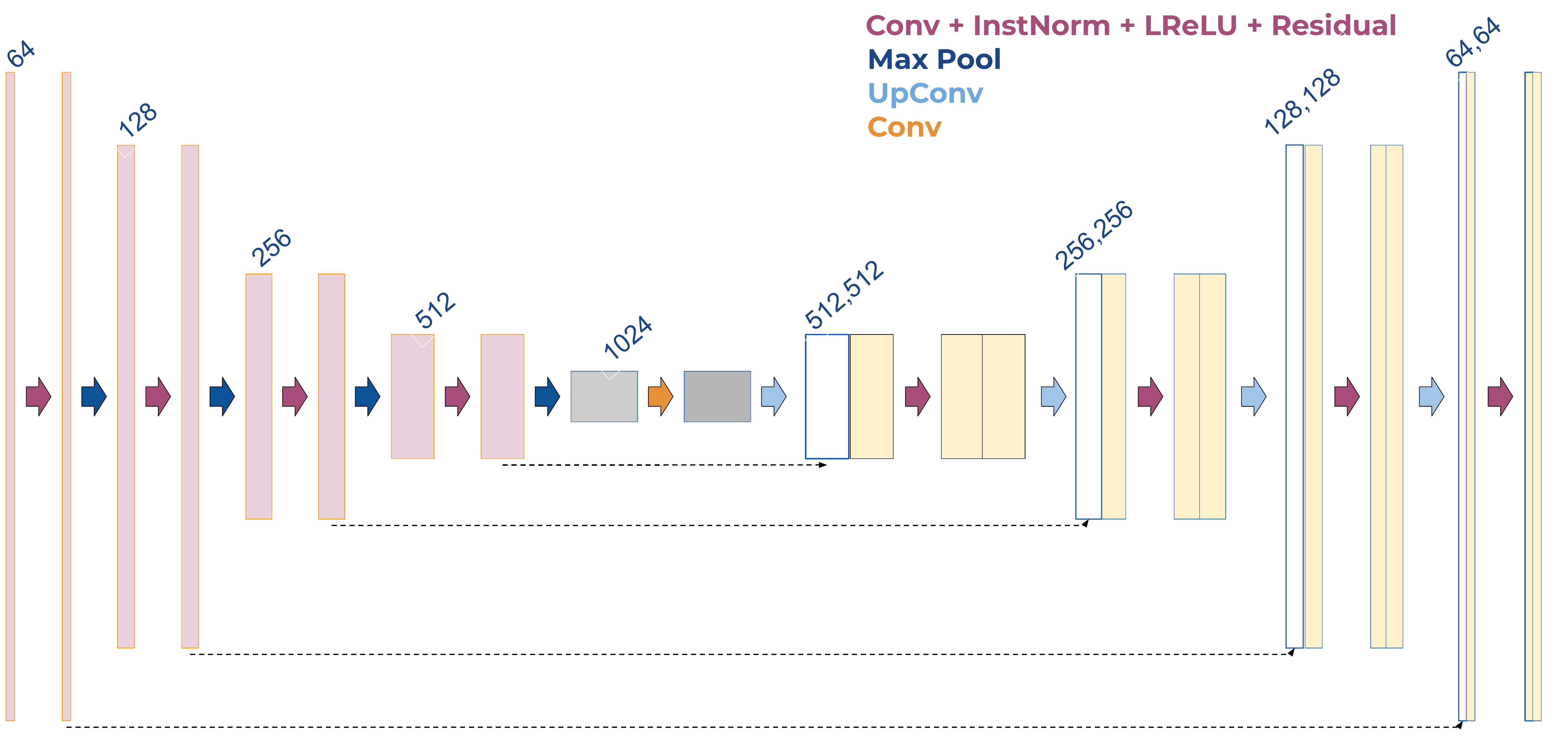}
    \caption{3D U-net architecture as provided by the FETS challenge.}
    \label{unet}
\end{figure}

The segmentation network is a 3D-Unet. It was provided by the challenge organizers and remained unchanged during all experiments. The architecture is composed of an encoder with residual branches followed by a decoder. We use the \emph{LeakyReLu} activation function \cite{maas2013rectifier} along with instance normalization \cite{ulyanov2016instance} - for mitigating the covariate shift. Dice serves as loss function. Fig. \ref{unet} illustrates the schematic of the network.

\subsection{Federated Cost Weighted Averaging (FedCostWAvg)}
The gold standard federated averaging (FedAvg) approach updates the global model as an average of all local models weighted by the respective sizes of the training data set. The new model $M_{i+1}$ is calculated as follows: \begin{equation}
    M_{i+1} = \frac{1}{S} \sum_{j=1}^n s_j M_{i}^j
\end{equation} where $s_j$ is the number of samples that model $M^j$ was trained on in round $i$ and $S = \sum_j s_j$. We propose a new weighting strategy that includes the amount by which the cost function decreased during the last step. Using FedCostWAvg, the new model $M_{i+1}$ is calculated as following: 
\begin{equation}
    M_{i+1} = \sum_{j=1}^n (\alpha \frac{s_j}{S} + (1 - \alpha) \frac{k_j}{K}) M_{i}^j 
\end{equation}
with:
\begin{equation}
     k_j = \frac{c(M_{i-1}^j)}{c(M_{i}^j)}, K = \sum_j k_j
\end{equation}

where $c(M_i^j)$ is the cost of the model $j$ at timestep $i$ that is simply calculated from the cost function that is being used to train the models locally. $\alpha$ is a parameter ranging between $0$ and $1$ that can be chosen to determine the balance between data size and cost improvement. In our experiments, a value of \emph{$\alpha = 0.5$} performed best. Intuitively, this weighting strategy adjusts not only for the training data set size but also for the size of the local improvements that were made during the last training round. Local updates which only marginally improved the local cost will influence the global update to a lesser extent than those which had a bigger impact.

\section{Results}
The method won the challenge and significantly outperformed all other submitted methods; tables \ref{tab:score} and \ref{tab:score2} summarize the performance upon convergence. \\

\begin{figure}[H]
    
    \centering
    \includegraphics[width=\textwidth]{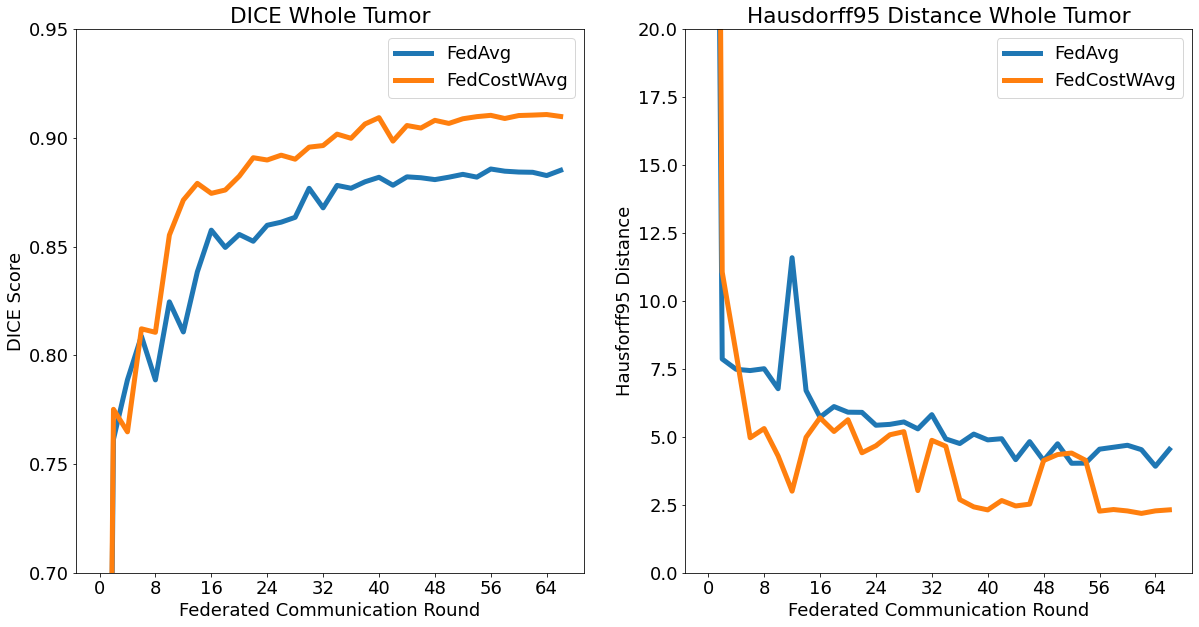}
    \caption{\label{wt} Comparison of the DICE Whole Tumor metric per federated round for FedCostWAvg vs. FedAvg. Note of course that the bigger the DICE score, the better and the smaller the Hausdorff95 distance, the better. }

\end{figure}

\begin{figure}[H]
    
    \centering
    \includegraphics[width=\textwidth]{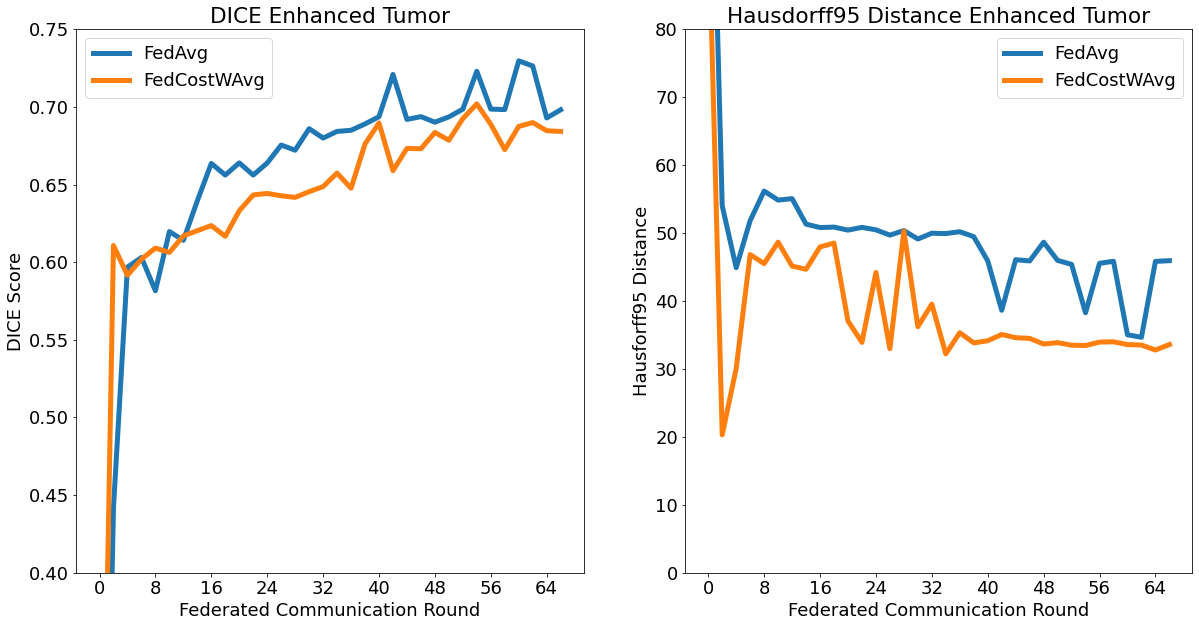}
    \caption{\label{et} Comparison of the DICE Enhanced Tumor metric per federated round for FedCostWAvg vs. FedAvg. Note of course that the bigger the DICE score, the better and the smaller the Hausdorff95 distance, the better. }

\end{figure}

\begin{figure}[H]
    
    \centering
    \includegraphics[width=\textwidth]{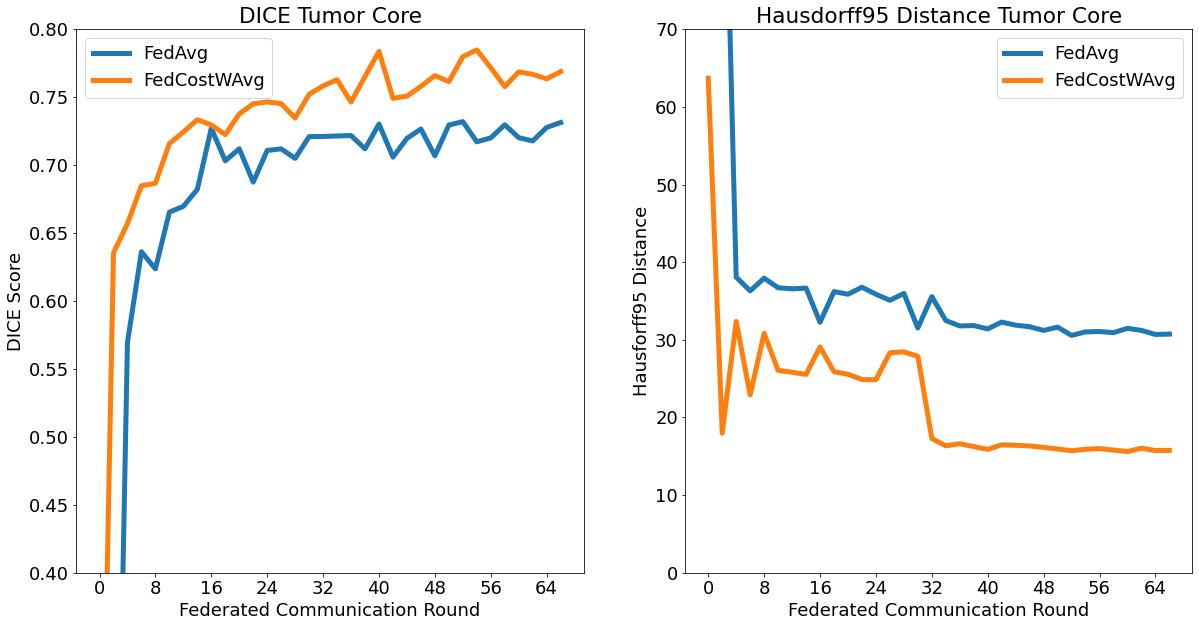}
    \caption{\label{tc} Comparison of the DICE Tumor Core metric per federated round for FedCostWAvg vs. FedAvg. Note of course that the bigger the DICE score, the better and the smaller the Hausdorff95 distance, the better. }

\end{figure}

In addition we used the provided data (which is a smaller subset of the challenge data) to test the performance of FedCostWAvg against FedAvg in order to visualize the convergence behaviour. We trained and validated the model on 369 samples which were unevenly distributed over 17 data centers. The training-validation split was $80/20$, the learning rate was $1e-4$ and we did $10$ epochs per federated round. Please note that computational resources were limited so no exhaustive grid search to find optimal hyperparameters was feasible, also training could not run long enough to achieve maximal performance. Figures \ref{wt}, \ref{et} and \ref{tc} depict the performances over communication rounds. Also note that of course the most informative comparison between methods was done in the challenge itself with more data and many different initialisations. This comparison serves only as a visualization of how different convergence behaviours look like for one initialisation. We observe an improvement for almost all classes and metrics, when using our proposed method. The exemption is the DICE Enhanced Tumor Metric. Note though that the difference is not significant and the methods have not yet converged. 

\subsection{Discussion}
While these results already show a clear improvement over FedAvg, it is unclear whether other hyperparameters would have achieved an even better result. Due to limitations in training resources a proper grid search was not feasible. \\

The simple and straightforward interpretation of the mechanism of FedCostWAvg is amplification of more informative updates against less informing ones. It could be seen as a diminishing returns acknowledging method. A deeper insight might be the interpretation as resembling a PID controller\footnote{The credit for this observation goes to David Naccache.} \cite{bellman2015adaptive}. When one reframes the federated learning problem as a control problem, then the central server that does the averaging is equivalent to a control unit that is included in a feedback loop. When one would then extend this logic to the averaging approach, it might be intelligent to view FedCostWAvg as an approximation of a PID controller, where the newly added term corresponding to the drop in cost is effectively functioning as the derivative part and the data size term as the proportional one. Future research could try to include the integral term as well.  

\begin{table}
\begin{adjustwidth}{0in}{0in}
    \centering
    \setlength\tabcolsep{3pt} 
    \begin{tabular}{|l|c|c|c|c|c|c|c|c|c|c|c|c|c|c|}
        \hline
        
        Label & DICE WT & DICE ET & DICE TC & Sens. WT & Sens. ET & Sens. TC   \\ \hline
        Mean & 0,8248 & 0,7476 & 0,7932 & 0,8957 & 0,8246 & 0,8269  \\ \hline 
        StdDev & 0,1849 & 0,2444 & 0,2643 & 0,1738 & 0,2598 & 0,2721  \\ \hline
        Median & 0,8936 & 0,8259 & 0,9014 & 0,948 & 0,9258 & 0,9422  \\ \hline
        25th quantile & 0,8116 & 0,7086 & 0,8046 & 0,9027 & 0,7975 & 0,8258 \\ \hline
        75th quantile & 0,9222 & 0,8909 & 0,942 & 0,9787 & 0,9772 & 0,9785  \\ \hline

    \end{tabular}
    \vspace{2pt}
    \caption{\label{tab:score} Final performance of FedCostWAvg in the FETS Challenge, DICE and Sensitivity}
\end{adjustwidth}
\end{table}

\begin{table}
\begin{adjustwidth}{0in}{0in}
    \centering
    \setlength\tabcolsep{3pt} 
    \begin{tabular}{|l|c|c|c|c|c|c|c|c|c|c|c|c|c|c|}
        \hline
        Label & Spec WT & Spec ET & Spec TC & H95 WT & H95 ET & H95 TC & Comm. Cost  \\ \hline
        Mean & 0,9981 & 0,9994 & 0,9994 & 11,618 & 27,2745 & 28,4825 & 0,723 \\ \hline
        StdDev & 0,0024 & 0,0011 & 0,0014 & 31,758 & 88,566 & 88,2921 & 0,723 \\ \hline
        Median & 0,9986 & 0,9996 & 0,9998 & 5 & 2,2361 & 3,0811 & 0,723 \\ \hline
        25th quantile & 0,9977 & 0,9993 & 0,9995 & 2,8284 & 1,4142 & 1,7856 & 0,723 \\ \hline
        75th quantile & 0,9994 & 0,9999 & 0,9999 & 8,6023 & 3,5628 & 7,0533 & 0,723 \\ \hline

    \end{tabular}
    \vspace{2pt}
    \caption{\label{tab:score2} Final performance of FedCostWAvg in the FETS Challenge, Specificity, Hausdorff95 Distance and Communication Cost}
\end{adjustwidth}
\end{table}

\section{Conclusion}
In this paper, we describe a method for model aggregation developed for the MICCAI Federated Tumor Segmentation Challenge (FETS). The novelty of the method lays in including local cost improvements when calculating the weights for averaging models which are trained at different centers. The approach is validated on a brain tumor segmentation task and achieves the best performance among all participating teams.

\section*{\large{Acknowledgements}}
\noindent Bjoern Menze, Benedikt Wiestler, and Florian Kofler are supported through the SFB 824, subproject B12. Supported by Deutsche Forschungsgemeinschaft (DFG) through TUM International Graduate School of Science and Engineering (IGSSE), GSC 81. Suprosanna Shit and Ivan Ezhov are supported by the Translational Brain Imaging Training Network (TRABIT) under the European Union's `Horizon 2020' research \& innovation program (Grant agreement ID: 765148). With the support of the Technical University of Munich – Institute for Advanced Study, funded by the German Excellence Initiative. Ivan Ezhov is also supported by the International Graduate School of Science and Engineering (IGSSE). Johannes C. Paetzold and Suprosanna Shit are supported by the Graduate School of Bioengineering, Technical University of Munich.

{ \bibliographystyle{splncs}
\bibliography{mybibliography}
}

\end{document}